\documentclass[letterpaper]{article}
\usepackage{flairs}%aaai
\usepackage{times}
\usepackage{helvet}
\usepackage{courier}
\usepackage{subfigure}
\usepackage{caption}

\usepackage{graphicx}
\usepackage{xcolor, soul}
\usepackage{csquotes}
\sethlcolor{pink}

\begin{document}
% This file is an adoption of the style file for AAAI Press 
% proceedings, working notes, and technical reports.  This file is made 
% with minimal changes by explicit permission from AAAI.
\title{Assessing the Impact of Sequence Length Learning on Classification Tasks for Transformer Encoder Models}
\author{Jean-Thomas Baillargeon, Luc Lamontagne\\
Departement d’informatique et de génie logiciel, Université Laval, QC, Canada\\
\{jean-thomas.baillargeon, luc.lamontagne\}@ift.ulaval.ca\\
}
\maketitle
\begin{abstract}
\begin{quote}
Classification algorithms using Transformer architectures can be affected by the sequence length learning problem whenever observations from different classes have a different length distribution. This problem causes models to use sequence length as a predictive feature instead of relying on important textual information. Although most public datasets are not affected by this problem, privately owned corpora for fields such as medicine and insurance may carry this data bias. The exploitation of this sequence length feature poses challenges throughout the value chain as these machine learning models can be used in critical applications. In this paper, we empirically expose this problem and present approaches to minimize its impacts.   
\end{quote}
\end{abstract}

\noindent  
\section{Introduction}
\label{intro}
Transformer-based models \cite{vaswani2017attention}, the current go-to models in terms of state-of-the-art performance, achieve impressive performances in most natural language processing (NLP) tasks. Being parameter-heavy models trained on huge corpora, they retain most of the training data information \cite{bender2021dangers}, leading to significant improvements. However, these improvements sometimes rely on unknown and undesirable correlations present in the training and test data. Models can learn to leverage these correlations as classification shortcuts \cite{bastings2021will}. 

A well-known source of shortcuts is bias. Using bias as a shortcut, a model could, for instance, associate gender and occupation together \cite{lu2020gender} instead of learning useful textual representations that would help infer someone's occupation. Even though the use of shortcuts can lead to model improvements during construction as they are good for the wrong reasons \cite{mccoy2019right}, it can create problems along the value chain if end users are not aware of the heuristic used. Other sources of classification shortcut include spurious features where inductive model bias is generated by artifacts from the annotation process or text structure, so models learn the dataset instead of the task.

A lesser-known classification shortcut is the sequence length difference between observations of different classes in a classification task. Exploiting this feature implies that any variation of the observations length, regardless of their textual content, can lead to misclassifications. As with other spurious features, the use of shortcuts also impacts the performance of explanation mechanisms since the main features exploited are not based on textual content. Another problem with sequence length learning is artificial overperformance with the training and test datasets. A performance distortion leads to unfair comparisons for model selection since some models may converge to a local minimum using sequence length rather than fully representing the textual information. Overall, sequence length should be avoided because the performance gain comes at the expense of model robustness, explainability, and evaluation. 

Although seldom encountered in public and open datasets, we encounter this problem in our work as we aim to provide analyst with early warnings of catastrophic (costly) claims. Since basic and catastrophic claim files have different length profiles, our modeling efforts are hampered by sequence length learning. 

This paper presents two contributions related to the problem described above. Our first contribution is an empirical study of how transformer-based models are affected by sequence length learning. Our second contribution is the evaluation of techniques using the capabilities of pretrained transformers to mitigate the impacts of this feature.

The remainder of the paper is as follows. We describe in Section \ref{sec:related_work} how the literature identifies and addresses the problem. In Section \ref{sec:evaluation_methodology}, we present a series of experiments that investigate the impact of the sequence length meta-feature using four textual classification datasets. Finally, we analyze in Section \ref{sec:simple_technique} data-centric techniques to mitigate the impact of learning sequence length meta-features.

\section{Related Work} \label{sec:related_work}

Model exploiting classification shortcuts is no new theme in the natural language processing research community. Many studies have been made regarding fairness and bias (such as gender or ethnicity, see \cite{garrido2021survey}) that bypass task learning by learning the artifacts from the dataset. \cite{lovering2021predicting} investigated how spurious features (e.g. lexical overlap) were preferred over the target feature (i.e. textual information) and correlated the extrability of a feature in a language model documents pretrained representation with its use in the classification task. 
 
In their work, \cite{warstadt2020learning} identified many spurious (surface) features and studied their impact on pretraining weights of transformer models such as BERT and RoBERTa. Those identified features were associated with nonlinguistic text singularities, such as word absolute position (sentence starts with "the"), sequence length (text is longer than $n$ words), lexical content (sentence contains "the"), word relative position ("the" is before "an") and orthography (is the sentence in title case). They concluded that the model would acquire a preference for linguistics features rather than surface features as long as enough examples were provided during self-supervised pretraining. Although they considered the sequence length as a surface feature, it has not been evaluated how strongly the model relies on the feature on a downstream task.

This lesser-known sequence length spurious feature has seldom been explored in the literature. \cite{baillargeon2022preventing} presented that recurrent neural networks would use the sequence length as a feature in an empirical demonstration using simulation and with a sentiment polarity task. They presented that weight decay regularization prevented the usage of sequence length. However, their work was limited to recurrent neural networks, and transformer architecture was left out. Our application could not eliminate the sequence length using their regularization scheme. Sequence length was also studied in \cite{jeon2021countering}, where the authors identified this potential problem in essay gradings. Their work presented that the transformer model (XLNet) uses the essay length to grade it in a regression task. To palliate this problem, the authors used the assumption that word distribution is a score predictor and is not transferable to our classification problem.

The literature on spurious features and bias mitigation proposed various solutions avenues, such as adversarial learning  \cite{belinkov2019adversarial} and optimizer-based solutions \cite{jiang2022rose}. Approaches, such as data augmentation, are presented by \cite{sun2019mitigating} and \cite{prost2019debiasing} to mitigate (gender) bias and reduce its impacts as a classification shortcut. \cite{wu2022generating} addressed the spurious feature problem with a data augmentation and reduction approach for datasets for natural language inference tasks. Both methods we are proposing are inspired by their approach.  

% Other authors, such as \cite{varivs2021sequence}, \cite{anil2022exploring} and \cite{press2021train}, considered this problem as an out-of-distribution (OOD) generalization problem for sequence-to-sequence models. Our work shows that transformer-based classifiers generalize very well to OOD examples and will not be addressed as such.

\section{Assessing the Impact of Sequence Length Learning}\label{sec:evaluation_methodology}
 In this section, we describe our approach to expose models to sequence length learning using publicly available classification datasets. We start by presenting the datasets used in our work and then describe various experiments to study this problem. In the first experiment, we assess the propensity of models to rely on sequence length. Other aspects, such as the extent of the problem resulting from overlapping class length distributions, are studied in subsequent experiments. We finally show that the problem affects different transformer encoder architectures and we explore its potential sources.

\subsection{Datasets}
In our experiment, we expose four text classification datasets to the sequence length learning problem. The first dataset, \textbf{Amazon-Polarity} (AP) \cite{zhang2015character}, contains reviews on a scale of five ratings grouped into negative (label 0) and positive (label 1) classes. The training set consists of 3.6M examples equally divided into two classes. The second, \textbf{Yelp-Polarity} (YP) \cite{zhang2015character}, is another binary classification dataset containing 500K text examples labeled negative (0) or positive (1). The third dataset is \textbf{Multi Natural Language Inference} (MNLI), introduced by \cite{williams2018broad}, a multi-class classification task included in the GLUE benchmark. This dataset contains 433k sentence pairs and three inference labels, where a model must assess whether the second sentence is either an entailment (label 0), neutral (label 1), or contradictory (label 2) with the first sentence. Finally the fourth dataset, Question-answering NLI (QLNI) introduced by \cite{rajpurkar2016squad}, also included in GLUE, is a binary classification with 116k question-answer pairs, labeled as entailment (0) or not-entailment (1).

To expose the datasets to the sequence length problem, we inject this spurious feature by creating a sequence length imbalance in the training data, and we partition the test set to assess the behavior of the model for different overlaps of distribution.

\subsubsection{Alteration of Training Datasets to Inject sequence length Imbalance}\label{subsubsec:training_subsets}
The sequence length meta-feature is injected by truncating the training sets to obtain non-overlapping sequence length distributions between classes. To do this, a specific length threshold is selected to divide the class observations. As we use transformers with specific pretrained tokenizers, we estimate the sequence length as the number of tokens after tokenizing the dataset. Also, as transformer models have a maximum input size, observations longer than this maximum size are truncated. For binary classification, we select the length threshold that gives the best accuracy score for a classifier that uses the sequence length of observations as the only feature. For the NLI tasks, we used the 33rd and 66th (for MNLI) and 50th (for QNLI) length percentile of the full training set distribution. Table \ref{tab:baseline_2} contains these values for the four datasets used in our work.

% \begin{table}[hbt]
%     \centering
% \begin{tabular}{ cccc }
% \hline
% Dataset & Short Class & Long Class & Overlap \\
%         &  (Avg. Len.) & (Avg. Len.) &  \\
% \hline
% AP & 1 (77) & 0 (88) &  92.4 \%  \\ 
% YP  & 1 (111) & 0 (145) & 88.0 \%  \\ 
% \hline
% \end{tabular}
% \vspace{0.2cm}
% \caption{Short and long classes for the two datasets.}
% \label{tab:baseline_1}
% \end{table}

% We find that the negative class examples tend to be longer in both datasets, with a median of 88 tokens for AP and 145 for YP. The positive observations of AP of YP have a median length of 77 and 111 tokens respectively. Both datasets have a healthy overlap percentage of 92.4 \% (AP) and 88 \% (YP). These overlap values suggest that sequence length learning should not be present in the original versions of the datasets. 

\begin{table}[hbt]
    \centering
\begin{tabular}{ cc }
\hline
Dataset & Threshold \\
        
\hline
AP &  92  \\ 
YP  &  127 \\ 
MLNI  &  31, 45  \\ 
QLNI  &  48  \\ 

\hline
\end{tabular}
\vspace{0.2cm}
\caption{Threshold lengths used for dataset partitioning.}
\label{tab:baseline_2}
\end{table}
As noted in Table \ref{tab:baseline_2}, a threshold of 92 tokens is used to partition the examples for Amazon Polarity and 127 tokens for Yelp Polarity. For both datasets, we keep the negative examples above the threshold and the positives below. % Although arbitrary, this criterion was chosen to respect the fact that both datasets had, on average, longer negative observations. 
For MNLI, we keep observations containing up to 31 tokens for entailment, 32 to 45 for neutral, and above 45 for contradiction. For QNLI, we kept observation containing up to 48 tokens for entailment and above 48 for not-entailment.

\subsubsection{Generation of Test Subsets}\label{subsubsec:testing_subsets}
To evaluate the importance of sequence length in  the classification process, we divide each test datasets according to two partitionings. The first test partitioning, \textbf{Gap-test}, uses the same rules as in Section \ref{subsubsec:training_subsets} to keep test length distributions similar to those in the training set. The resulting partitions contain task-relevant data and are also of unbalanced length. The second, \textbf{Reverse-test}, is partitioned as the complement of the first by filtering observations (instead of keeping them) using the threshold rules. This second partitioning still contains relevant data for the task. However the length distributions in these test partitions are the inverse of those in the training set. In Figure \ref{fig:yelp_test_distribution}, we present the partitionings obtained for Yelp Polarity. Observing a difference in the performance of a model on these two datasets would indicate wether the model uses sequence length as spurious feature.

\begin{figure} [hbt!]
  \centering

   \captionsetup{labelformat=empty}
    \centering\includegraphics[width=0.35\textwidth]{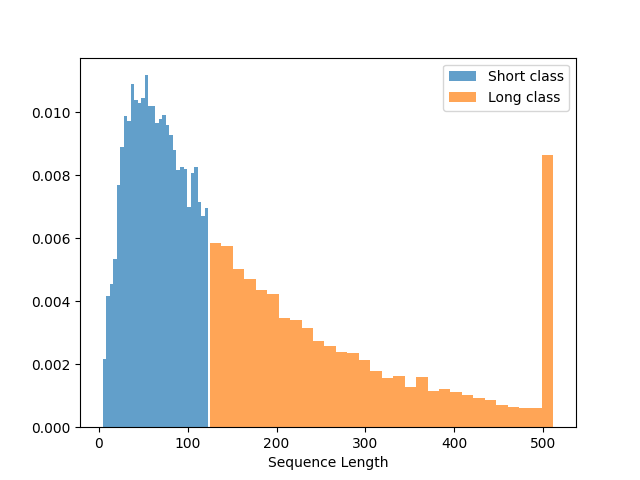}
    \caption{(a) Gap-test partitioning}

    \centering\includegraphics[width=0.35\textwidth]{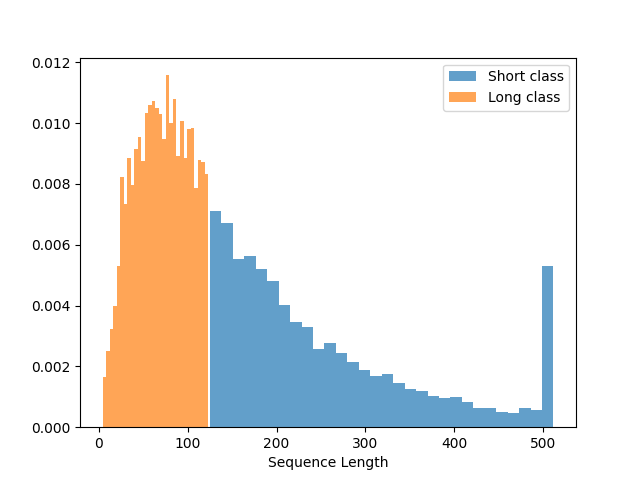}
    \caption{(b) Reverse-test partitioning}
    \setcounter{figure}{1}    
    \captionsetup{labelformat=original}
\caption{The length distributions for different partitionings of the Yelp Polarity test set. The spikes on the right side correspond to the proportion of examples truncated at the maximum transformer input length.}  
\label{fig:yelp_test_distribution}
   
  \end{figure}

\subsection{Evaluation of the Impact of the sequence length Feature} \label{subsec:eval_fullimpact}
To assess the presence of sequence length learning, we train models on the training datasets altered as described in Section \ref{subsubsec:training_subsets} and evaluate them on the three test sets (the \textbf{Original} test set as well as its \textbf{Gap-test} and \textbf{Reverse-test} partitionings) presented in Section \ref{subsubsec:testing_subsets}. 
 For this experiment, we use {\tt RoBERTa} \cite{liu2019roberta}, a baseline model that we finetune from the {\tt roberta-base} pretrained weights provided by the Hugging Face transformers library \cite{wolf2019huggingface}. % Model hyperparameters and topology are described in the appendix. 

\begin{table}[hbt!]
\centering
\begin{tabular}{ cccc }
 \hline
Original &  Original & Gap & Reverse \\
train & test& test & test \\
\hline
AP & 94.7 \%  & 95.1 \%  & 94.1 \%  \\
YP & 97.6 \% & 98.0 \% & 97.1 \%  \\
MNLI & 82.9 \% & 82.1 \% & 83.3 \%  \\
QLNI & 91.4 \% & 91.3 \% & 91.6 \%  \\
\hline

\end{tabular}
\vspace{0.2cm}
\caption{Single-run accuracy with the original train sets}
\label{tab:baseline_3}
\end{table}

We first present in Table \ref{tab:baseline_3} the classification results of the models trained with the original unmodified training datasets. We can observe that the accuracy performance is similar to the state of the art for each dataset. Moreover using length-unbalanced test sets does not affect the results because classification the models are trained on length-balanced training sets. 
 
\begin{table}[hbt!]
\centering
\begin{tabular}{ cccc }
 \hline
Altered &  Original & Gap & Reverse \\
train & test& test & test \\
\hline
AP &  53.4 \% & 100.0 \% & 0.0 \%  \\
YP &  56.0 \% & 100.0 \% & 0.0 \%  \\
MNLI & 34.1 \% & 100.0 \% & 13.3 \%  \\
QNLI & 46.3 \% & 100.0 \% & 0.0 \%  \\

\end{tabular}
\vspace{0.2cm}
\caption{Single-run accuracy with the altered datasets}
\label{tab:baseline_4}
\end{table}

To evaluate the impact of unbalanced sequence length on model behavior, we finetune other {\tt roberta-base} models using the length-altered training sets described in Section \ref{subsubsec:training_subsets}. Table \ref{tab:baseline_4} presents the evaluation results with these models.

First we can see that all four models perform very well on the \textbf{Gap-Test} partitioning with accuracy results above the baseline evaluation. In fact, all four obtain perfect scores. This first result suggests that all the models seem to capture sequence length as a classification spurious feature. 

However, we observe that all the models perform very poorly on the \textbf{Reverse-Test} partitioning. Both sentiment analysis and QNLI models fail to make a single accurate prediction. Moreover, the results of the MNLI model are significantly lower than those of a random class selection. Again, it suggests that the models rely heavily on sequence length and also indicates that the content of the texts is almost never taken into account to make predictions.

A last observation made from the results obtained on the \textbf{Original-Test} partitioning indicates that training the models with length-unbalanced datasets significantly degrades performance even when evaluated with a balanced test set.    

Combining all these observations, we can draw two conclusions with a high degree of certainty. The first one is that the model will strongly rely on sequence length to make classifications if the data for each class is non-overlapping partitioned. The second conclusion is that the performance of a transformer model will be deceptively high when sequence length can be used. 
%Another perspective of this problem is that both the altered training dataset and the \textbf{Gap-Test} set are drawn from the same data distribution, thus fulfilling an important assumption of machine learning methodologies. 
Furthermore, there is a high risk that analysts would be unaware of the presence of the sequence length feature contribution to the models and would be unable to evaluate its significance.

\subsection{Evaluation of sequence length Learning for Partial Class Overlap} \label{sec:eval_overlapping}

We concluded in Section \ref{subsec:eval_fullimpact} that sequence length learning will bring the model to exclusively use this spurious feature when the length distribution of the different classes is perfectly partitioned. However, in a more realistic scenario, the classes will have overlapping sequence length distributions, and its impact will not be as drastic as presented in the previous section. In this section, we evaluate the relationship between the percentage of overlap in the length distributions of two classes and its impact on the classification performance.

To control the injection of the sequence length meta-feature in the datasets, we gradually adjust the overlap ratio between the two classes using different lower and upper thresholds. Table \ref{tab:lower_upper} presents the upper and lower bounds of the selected overlap proportions for the Amazon Polarity and Yelp Polarity classes. 

\begin{table}[hbt]
\begin{tabular}{ cccccc }
\hline
Dataset & Overlap \% & Lower & Upper \\
\hline
AP & 92 \% (original) & 0  & 512 \\
AP & 80 \% &  40 & 200 \\
AP & 50 \% &  60 & 125 \\
AP & 25 \% &  70 & 85 \\
AP & 0 \% & 80  & 80 \\

\hline
YP & 87 \% (original) & 0 & 512 \\
YP & 80 \% & 30  & 360 \\
\textbf{YP} & \textbf{50} \% &\textbf{ 90}  & \textbf{230} \\
YP & 25 \% & 100  & 150 \\
YP & 0 \% & 125  & 125 \\
\hline
\end{tabular}
\vspace{0.2cm}
\caption{Lower and upper thresholds for partitioning examples respectively with long and short lengths}
\label{tab:lower_upper}
\end{table}

For illustration purposes, the reader can understand from the values in bold in the table that if we keep examples with less than 230 tokens for the short class and examples longer than 90 tokens for the long class of YP, one obtains two distributions where 50\% of the examples overlap in sequence length. This partitioning of the dataset is illustrated in Figure \ref{fig:yelp_distribution} with the original distribution (88\% overlap) in part (a) and the altered version in part (b).

\begin{figure}[htb!]
  \centering

\captionsetup{labelformat=empty}
    \centering\includegraphics[width=0.35\textwidth]{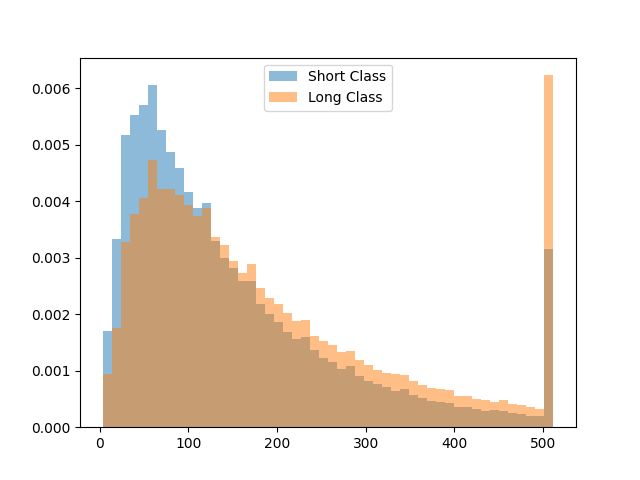}
    
    \caption{(a) Original distribution}

    \centering\includegraphics[width=0.35\textwidth]{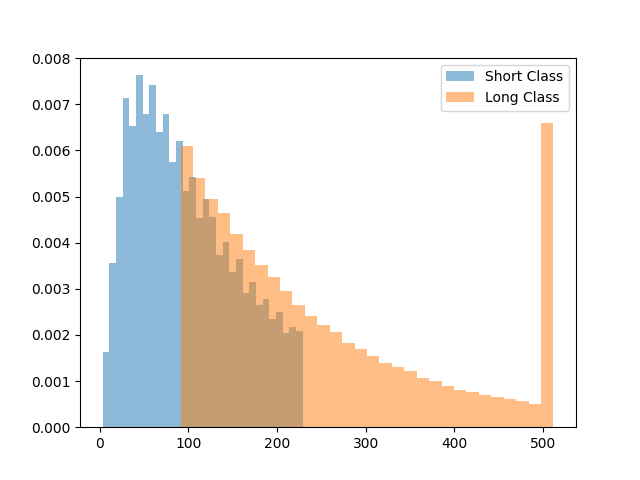}
    \caption{(b) Altered distribution with 50 \% overlap}
    \captionsetup{labelformat=original}
    \setcounter{figure}{0}    
  \caption{Observation length distributions for Yelp-Polarity.}  
    \label{fig:yelp_distribution}
  \end{figure}
\begin{table}[htb!]
\centering
\begin{tabular}{ ccccc }
 \hline
Altered & Overlap & Original & Gap & Reverse  \\
train & \% & test & test & test \\
\hline
AP & 92 \% & 95 \% & 95 \% & 94 \%  \\
AP & 80 \% & 88 \% & 97 \% & 77 \%  \\
AP & 50 \% & 70 \% & 98 \% & 38 \%  \\
AP & 25 \% & 60 \% & 99 \% & 15 \%  \\
AP & 0 \% & 53 \% & 100 \% & 0 \%  \\
\hline
YP & 88 \% & 98 \% & 98 \% & 97 \%  \\
YP & 80 \% & 94 \% & 99 \% & 89 \%  \\
YP & 50 \% & 75 \% & 99 \% & 43 \%  \\
YP & 25 \% & 64 \% & 100 \% & 19 \%  \\
YP & 0 \% & 56 \% & 100 \% & 0 \%  \\
\hline
\end{tabular}
\vspace{0.2cm}
\caption{Single-run accuracy of models trained with datasets altered for different overlap ratios.}
\label{tab:baseline_results}
\end{table}
From the results presented in Table \ref{tab:baseline_results}, we first notice that, as the training distribution overlap diminishes, the performance on \textbf{Original-test} degrades for both datasets, from 95.0 \% to 53.2 \% for AP and from 97.6 \% to 56.0 \% for YP. As \textbf{Original-test} combines the observations from \textbf{Gap-test} and \textbf{Reverse-test}, we expect this decrease to come from either partition. Investigating the performance of the models on \textbf{Gap-test} allow us to observe an increase in accuracy as the overlap proportion of both classes decreases. Conversely, the accuracy of the models evaluated on the \textbf{Reverse-test} subset decreases substantially, with a degradation almost linearly proportional to the overlap ratio. This suggests that a sequence learning effect takes place and becomes stronger as the percentage of overlap decreases (corresponding to an increase of the sequence length imbalance).

Using this analysis, we conclude that the transformer encoder model used for these experiments suffer from the sequence length learning problem to a large extent whenever the spurious surface feature is present in the dataset. The overlap ratio between the class length distributions offers a simple and efficient estimator of the extent of the problem in a dataset. The more the distributions overlap, the lesser the problem. 

\subsection{Source of Sequence Length Learning in Transformers Layers}
An important question to ask is "Where does this learning in the transformer architecture come from?". To locate the source of the problem, we divide {\tt RoBERTa} into three parts (embedding layer, encoder layers and classification head) and train each part in isolation using the altered training dataset as in Section \ref{subsec:eval_fullimpact}. We then assess the extent to which sequence length learning occurs in these trained parts. We trained each part for 10 epochs.

The first part is the \textbf{embedding} layer that contains the embeddings of the tokens, the positions and the token types. The second part contain the transformer layers of the \textbf{encoder}, composed (for {\tt RoBERTa}) of 12 transformer blocs featuring the self-attention mechanism. Finally, the last layer is the \textbf{classification} head, which includes a fully connected layer that converts the {\tt <cls>} token generated by the encoder into an output class prediction.

\begin{table}[htb!]
\centering
\begin{tabular}{ cccc }
 \hline
Trained & original & gap & reverse  \\
Bloc  & test & test & test \\

\hline

\hline
Embedding & 55 \% &  54 \% & 55 \%  \\
Encoder &  56 \% & 100 \% & 0 \%  \\
Classification & 77 \% & 98 \% & 51 \%  \\

\end{tabular}
\vspace{0.2cm}
\caption{Single-run evaluation (accuracy) of each transformer part trained in isolation.}
\label{tab:results_by_blocs}
\end{table}

We observe from Table \ref{tab:results_by_blocs} that the sequence length spurious feature strongly affects the transformer encoder layers and is also present to a lesser extent in the fully connected classification head. Surprisingly, no learning takes place in the Embedding layers, although positional embeddings would seem to be candidates for capturing sequence length related information.

\subsection{Sequence Length Learning for Different Transformer Encoder Architectures}

Our previous experiments are intended to illustrate the extent of the sequence length learning problem using a single model, {\tt roberta-base}. We present in this section the results for three other transformer encoder architectures. Those architectures were selected to encompass a wider range of models and extend our findings to the family of transformer encoders. 

 {\tt RoBERTa} was initially selected due to its high performance, availability, and general public adoption. Here we compare the {\tt base} and {\tt large} versions of RoBERTa to understand the impact of a large number of parameters on sequence length learning. {\tt Electra} was selected to evaluate a model not pretrained with Masked Language Modeling. Finally, {\tt BigBird} was chosen as a representative of the {\tt TransformerXL} family, which can extend the transformer architecture to sequences of longer lengths. This empirical study was performed using the YP dataset, and the models were trained for one full epoch, at which point they converged.

\begin{table}[!htb]
\centering
\begin{tabular}{ ccccc }
 \hline
Transformer & original & gap & reverse  \\
Model & test & test & test \\
\hline
RoBERTa-base & 56.4 \% & 100 \% &  0.001 \%  \\ % 0.21 epoch
RoBERTa-large & 56.7 \% & 100 \% &  0.005 \%  \\ % 0.38 epoch
Electra & 57.6 \% & 100 \% &  0.023 \%  \\
BigBird & 56.9 \% & 100 \% &  0.016 \%  \\
\end{tabular}
\vspace{0.2cm}
\caption{Single-run evaluation (accuracy) for different transformer encoder architectures trained with sequence length imbalanced data.}
\label{tab:transformer_architecture}
\end{table}

Results presented in Table \ref{tab:transformer_architecture} clearly indicate that all the models studied are affected and rely almost solely on sequence length class imbalance whenever this feature is present in the dataset.

\section{Alleviating the Impact of Sequence Length Learning }\label{sec:simple_technique}

Sequence length learning can be addressed as a bias or spurious feature learning problem. In both cases, as presented in Section \ref{sec:related_work}, solutions including adversarial learning or gradient adjustment could be good candidates. In this section, we favor data-oriented techniques that have proven to be effective by many authors, even if they are on the simpler side of the solution spectrum. We adopt two approaches to reduce the unwanted impact of sequence length learning on binary sentiment analysis models. The first is removing problematic observations from the training dataset and leveraging pretrained transformer representations. The second is a data augmentation technique that uses the transformer as a pretrained language model to increase the overlap percentage.

\subsection{Removing Problematic Observations}\label{ss:few-shot}
The first approach exploits the transformer representations learned during pretraining. These efficient representations allow us to significantly reduce the number of training examples by removing the observations that would create the greatest bias in the classification model. A similar scheme was adopted in \cite{wu2022generating}. In our work, we can trivially identify these observations by measuring their sequence length during data preparation. 

%As argued in \citet{brown2020language}, we believe that pre-trained models such as {\tt RoBERTa} are strong few-shot learners.

To evaluate this technique on binary sentiment analysis models, we finetune the weights of {\tt roberta-base} after removing all observations with a sequence length outside the overlap region of both class distributions. This results in the elimination of the presence of the sequence length meta-feature in training sets. %For instance, retaining long examples containing more than 70 tokens and short examples having less than 85 tokens, we obtain a training subset of AP reduced to 25\% of its original size with a class overlap of 100\%.

The results of the classification model evaluated on reduced datasets are presented in Table \ref{tab:reducing}. It contains the dataset used and the initial and updated class overlap percentage. The last two columns present the accuracies of the models trained with (Ini.) and without (Red.) problematic observations, evaluated on the unaltered \textbf{Original-test} dataset. 
\begin{table}[!htb]
\centering
\begin{tabular}{ ccccc }
 \hline
Train &  \multicolumn{2}{c}{Overlap}  & \multicolumn{2}{c}{Accuracy}  \\
Dataset  & Ini. & Red. & Ini. & Red. \\
\hline
AP & 80 \% & 100 \% & 87.7 \% & \textbf{95.4} \%  \\
AP & 50 \%& 100 \% & 70.3 \% & \textbf{95.2} \% \\
AP & 25 \%& 100 \% & 60.0 \% & \textbf{94.7} \% \\
AP & 0 \% & n/a  & 53.2 \% & n/a  \\
\hline
	
YP & 80 \%& 100 \% & 94.4 \% & \textbf{98.1} \%\\
YP & 50 \%& 100 \% & 74.6 \% & \textbf{97.8} \% \\
YP & 25 \%& 100 \% & 64.5 \% & \textbf{97.5} \%  \\
YP & 0 \% & n/a & 56.0 \% & n/a  \\
\hline
\end{tabular}
\vspace{0.2cm}
\caption{Single-run evaluation of models pre-trained with (Ini.) and without (Red.) problematic observations. The test set used is the \textbf{Original-test} dataset.}
\label{tab:reducing}
\end{table}

As expected, the models pre-trained without problematic examples perform very well. Even the models using a small part of the training data achieve performance similar to state-of-the-art models. % One should note that we cannot assess performance with a 0 \% overlap dataset as no observations are available for training purposes. 
% These results also show that OOD observations, with respect to sequence length, are classified with high accuracy. %This contrasts with results obtained on seq2seq tasks as discussed in Section \ref{sec:related_work}.

\subsection{Augmenting Training Data Using LM}
The second technique we tested exploits the transformer pre-trained language model to reduce the sequence length imbalance contained in the training data. We reduce this bias by synthetically increasing the overlap percentage. To achieve this, we extend the short class examples and shorten the long class examples using mask tokens. For this experiment, we use the masked language model (MLM) of RoBERTa to achieve both alterations.

\subsubsection{Data Augmention Approach}
To extend examples of the dataset, we choose examples from the short class, and add a random number of {\tt <mask>} literal tokens in each instance. In our experiment, the random number of {\tt <mask>} insertions for a given document follows a binomial distribution with parameters $q=0.15$ and $m$ being equal to the number of tokens in the example. %The parameter $q$ is selected to replicate the training parameters of the RoBERTa base model. 

To reduce examples of the dataset, we pick examples from the long class and select a random number of tokens to remove. Instead of removing the tokens and running the risk of creating dissociated sentence fragments, we replace two consecutive words with a single {\tt <mask>} token, allowing the MLM to fill in the blank and connect the two sentence pieces. 

For both operations, the words replacing the {\tt <mask>} tokens are selected by the MLM transformer model. The augmented dataset contains the examples resulting from the two operations.

\subsubsection{Results with Augmented Training data}
Knowing that labels are mostly preserved during the augmentation process, we evaluate whether training models with augmented datasets can alleviate the impact of sequence length learning. The results are presented in Table \ref{tab:augmented_results}. It contains the dataset used and the distribution overlap values of the initial (Ini.) and augmented (Aug.) training sets. The last two columns present the accuracy of the models trained using the initial dataset or augmented data, the performance being evaluated with the unaltered \textbf{Original-test} dataset.

\begin{table}[hbt!]
\centering
\begin{tabular}{ ccccccc }
 \hline
Train  &  \multicolumn{2}{c}{Overlap} & \multicolumn{2}{c}{Accuracy}  \\
Dataset &  Ini. & Aug. & Ini. & Aug. \\
 
\hline
AP & 80 \% & 84 \% & 87.7 \% & \textbf{90.4} \%  \\
AP & 50 \% & 65 \% & 70.3 \% & \textbf{77.2} \%  \\
AP & 25 \% & 46 \% & 60.0 \% & \textbf{69.2} \%  \\
AP & 0 \% & 24 \% & 53.2 \% & \textbf{62.8} \%  \\
\hline

YP & 80 \% & 89 \% & \textbf{94.4} \% & 86.6 \% \\
YP & 50 \% & 62 \% & \textbf{74.6} \% & 73.9 \%  \\
YP & 25 \% & 40 \% & 64.5 \% & \textbf{79.1} \%  \\
YP & 0 \% & 15 \% & 56.0 \% & \textbf{69.0} \%  \\
\hline
\end{tabular}
\vspace{0.2cm}
\caption{Single-run accuracy of the models trained with (Aug.) or without (Ini.) augmented datasets. The test set used is the \textbf{Original-test} dataset.}
\label{tab:augmented_results}
\end{table}

We consider that data augmentation helps whenever we measure an improvement of accuracy with the augmented model over the original version and augmenting datasets at lower overlap values (0\%, 25\%, and 50\%) lessen the spurious feature impact for both AP and YP. 

The results from the last two sections support our hypothesis that data-centric approaches, such as extension and reduction, can alleviate the impacts of sequence length learning when they increase the overlapping percentage of class length distribution.

\section{Conclusion \& Future Works}

In this paper, we illustrate with some tests that transformer-based architectures suffer from sequence length learning. When trained with an unbalanced dataset, a model learns to use the difference between class length distributions instead of relying on important textual features. We empirically demonstrate that we can inject a sequence length meta-feature into a dataset and force a transformer-based classifier to use it.  We evaluated this problem on different NLP tasks and transformer encoder models. Finally, we show its impact can be reduced by eliminating examples outside of the overlap region or by augmenting training datasets using the MLM capabilities of transformers.

For future work, we would like to perform a similar evaluation on hierarchical architectures such as HAN \cite{yang2016hierarchical} and Recurrence/Transformer over BERT (roBERT by \cite{pappagari2019hierarchical}). We believe these models would also suffer from the same problem. However the hierarchical topology of these networks would allow for innovative approaches. Adversarial approaches could also be evaluated, such as training a model to classify while preventing another model from using its representation to predict the sequence length. %Finally, we would like to evaluate if other tasks, such as regression, are as impacted as classification.

% \section*{Limitations}
% The main limitation of this work is that the proposed solution is dataset and task sensitive. Even if the weights of a transformer can be adapted to a fundamental task such as sentiment classification, other classification tasks using a field-specific  vocabulary have to be evaluated.  Depending on the corpus, one may need to elect either technique and select an MLM replacement hypothesis.

% We believe the nature of the sentiment analysis task is important for the good performance of the models in certain sections of the paper. Models were pre-trained using general-purpose corpora containing an important proportion of words with positive and negative polarity. Therefore, by design, the model's pre-trained weights can be combined efficiently to generate useful representations for sentiment classification. 

% As presented, the technique does not fully eliminate the impact of the sequence length learning problem but rather reduces it. Further analysis will be required to evaluate if the effect of the proposed techniques is sufficient for other use cases.

% All references should be stored in the file "references.bib"
% Please do not modify anything below this line.
\bibliographystyle{flairs}
\bibliography{references}

\end{document}